\documentclass[letterpaper, 10 pt, conference]{ieeeconf}  

\IEEEoverridecommandlockouts                              

\usepackage{times} 
\usepackage[pdftex]{graphicx}
\usepackage[usenames,dvips,pdftex]{color}
\usepackage{wrapfig}
\usepackage{amsmath, amsfonts}
\usepackage{amssymb}
\usepackage{longtable}
\usepackage{wrapfig}
\usepackage{cite}
\usepackage{units}
\usepackage{url}
\usepackage{algorithm}
\usepackage[caption=true]{subfig}
\usepackage{pdfpages}
\usepackage{epstopdf}
\usepackage{caption}
\usepackage[noend]{algpseudocode}
\usepackage{array,multirow}
\usepackage{hhline}
\usepackage{booktabs}
\usepackage{stackengine}
\usepackage[percent]{overpic}
\usepackage{relsize}

\newcommand{\matr}[1]{\boldsymbol{#1}}	
\newcommand{\set}[1]{\boldsymbol{#1}}	

\newcommand{\domain}[1]{\mathbb{#1}}
\newcommand{\Epsilon}{\mathcal{E}}

\newcommand{\meqref}[2]{(\ref{#1}, \ref{#2})}

\makeatletter
\ifdefined \r@draft \newcommand{\cblue}[1]{\textcolor{blue}{#1}} 
\else \newcommand{\cblue}[1]{\iffalse{#1}\fi}
\fi
\makeatother

\newcommand{\Figure}[1]{Fig.~\ref{#1}}

\newcommand{\Table}[1]{Table~\ref{#1}}
\newcommand{\Section}[1]{Sec.~\ref{#1}}

\usepackage{xcolor}
\usepackage{cancel}

\usepackage{tabularx}
\usepackage[export]{adjustbox}

\title{\LARGE \bf
CNN for IMU Assisted Odometry Estimation using Velodyne LiDAR}

\author{Martin Velas, Michal Spanel, Michal Hradis, and Adam Herout
\thanks{All authors are with with the Faculty of Information Technology, Brno University of Technology, Czech Republic {\tt\small \{ivelas$\vert$spanel$\vert$ihradis$\vert$herout\} at fit.vutbr.cz}}
\thanks{*This work has been supported by the Artemis JU grant agreement ALMARVI (no. 621439), the TACR Competence Centres project V3C (no. TE01020415), and the IT4IXS – IT4Innovations Excellence project (LQ1602).}
}

\begin{document}

\maketitle
\thispagestyle{empty}
\pagestyle{empty}

\begin{abstract}
We introduce a novel method for odometry estimation using convolutional neural networks from 3D LiDAR scans. The original sparse data are encoded into 2D matrices for the training of proposed networks and for the prediction. Our networks show significantly better precision in the estimation of translational motion parameters comparing with state of the art method LOAM, while achieving real-time performance. Together with IMU support, high quality odometry estimation and LiDAR data registration is realized. Moreover, we propose alternative CNNs trained for the prediction of rotational motion parameters while achieving results also comparable with state of the art. The proposed method can replace wheel encoders in odometry estimation or supplement missing GPS data, when the GNSS signal absents (e.g. during the indoor mapping). Our solution brings real-time performance and precision which are useful to provide online preview of the mapping results and verification of the map completeness in real time.
%
%
%
%
\end{abstract}

\section{Introduction}

Recently, many solutions for indoor and outdoor \emph{3D mapping} using LiDAR sensors have been introduced, proving that the problem of \emph{odometry estimation} and \emph{point cloud registration} is relevant and solutions are demanded. The Leica\footnote{\url{http://leica-geosystems.com}} company introduced Pegasus backpack equipped with multiple Velodyne LiDARs, RGB cameras, including IMU and GNSS sensors supporting the point cloud alignment. Geoslam\footnote{\url{https://geoslam.com}} uses simple rangefinder accompanied with IMU unit in their hand-helded mapping products ZEB1 and ZEB-REVO. Companies like LiDARUSA and RIEGL\footnote{\url{https://www.lidarusa.com}, \url{http://www.riegl.com}} build their LiDAR systems primarily targeting outdoor ground and aerial mapping. Such systems require readings from IMU and GNSS sensors in order to align captured LiDAR point clouds. These requirements restrict the systems to be used for mapping the areas where GNSS sensors are available.

Another  common property of these systems is offline processing of the recorded data for building the accurate $3$D maps. The operator is unable to verify whether the whole environment (building, park, forest, \ldots) is correctly captured and whether there are no parts missing. This is a significant disadvantage, since the repetition of the measurement process can be expensive and time demanding. Although the orientation can be estimated online and quite robustly by the IMU unit, \emph{precise position information} requires reliable GPS signal readings including the online corrections (differential GPS, RTK, ...). Since these requirements are not met in many scenarios (indoor scenes, forests, tunnels, mining sites, etc.), the less accurate methods, like odometry estimation from wheel platform encoders, are commonly used.

We propose an alternative solution -- a frame to frame \emph{odometry estimation} using \emph{convolutional neural networks} from LiDAR point clouds. Similar deployments of CNNs has already proved to be successful in ground segmentation \cite{cnn-gseg} and also in vehicle detection \cite{cnn-vdet} in sparse LiDAR data.

The main contribution of our work is fast, real-time and precise estimation of positional motion parameters (translation) outperforming the state-of-the-art results. We also propose alternative networks for full 6DoF visual odometry estimation (including rotation) with results comparable to the state of the art. Our deployment of convolutional neural networks for odometry estimation, together with existing methods for object detection \cite{cnn-vdet} or segmentation \cite{cnn-gseg} also illustrates general usability of CNNs for this type of \emph{sparse LiDAR data}.

\begin{figure}
	\centering
    \includegraphics[width=\linewidth]{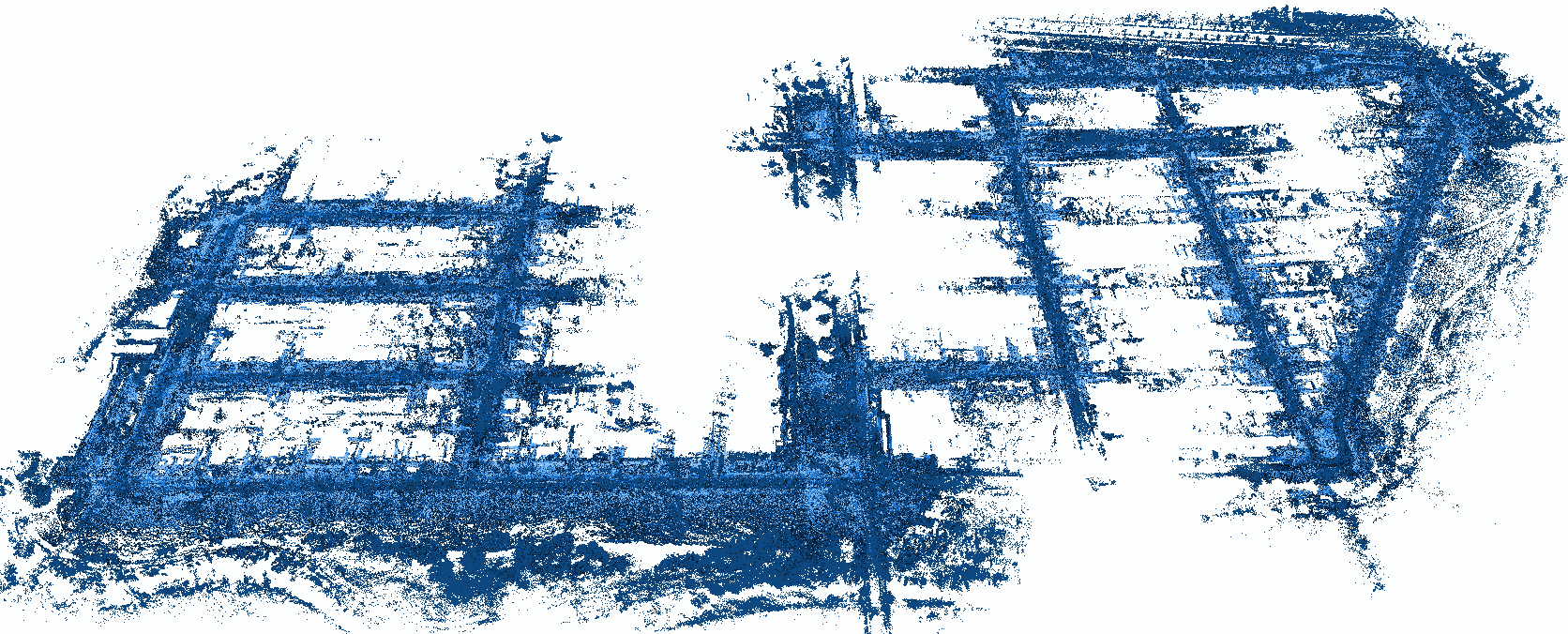}
    \caption{Example of LiDAR point clouds registered by CNN estimated odometry. Sequence $08$ of KITTI dataset \cite{kitti} is presented with rotations provided by IMU.\label{fig:teaser}}
    \vspace{-1em}
\end{figure}

\section{Related Work}
\label{sec:related-work}

The published methods for visual odometry estimation can be divided into two groups. The first one consists of direct methods computing the motion parameters in a single step (from image, depth or $3$D data). Comparing with the second group of iterative methods, direct methods have a potential of better time performance. Unfortunately, to our best knowledge, no direct method for odometry estimation from LiDAR data have been introduced so far.

Since the introduction of notoriously known Iterative Closest Point (ICP) algorithm \cite{Chen92,Besl92}, many modifications of this approach were developed. In all derivatives, two basic steps are iteratively repeated until the termination conditions are met: matching the elements between 2 point clouds (originally the points were used directly) and the estimation of target frame transformation, minimizing the error represented by the distance of matching elements. This approach assumes that there actually exist matching elements in the target cloud for a significant amount of basic elements in the source point cloud. However, such assumption does not often hold for sparse LiDAR data and causes significant inaccuracies.

Grant \cite{Grant13} used planes detected in Velodyne LiDAR data as the basic elements. The planes are identified by analysis of depth gradients within readings from a single laser beam and then by accumulating in a modified Hough space. The detected planes are matched and the optimal transformation is found using previously published method \cite{Pathak10}. Their evaluation shows the significant error ($\approx 1$m after $15$m run) when mapping indoor office environment. Douillard et al. \cite{Douillard12} used the ground removal and clustering remaining points into the segments. The transformation estimated from matching the segments is only approximate and it is probably compromised by using quite coarse ($20$cm) voxel grid.

Generalized ICP (GICP) \cite{gicp} replaces the standard point-to-point matching by the plane-to-plane strategy. Small local surfaces are estimated and their covariance matrices are used for their matching. When using  Velodyne LiDAR data, the authors achieved $\pm20$\,cm accuracy in the registration of pairwise scans. In our evaluation \cite{Velas16} using KITTI dataset \cite{kitti}, the method yields average error $11.5$cm in the frame-to-frame registration task. The robustness of GICP drops in case of large distance between the scans ($>6$m). This was improved by employing visual SIFT features extracted from omnidirectional Ladybug camera \cite{Pandey11} and the codebook quantization of extracted features for building sparse histogram and maximization of mutual information \cite{Pandey12}.

Bose and Zlot \cite{Bosse13} are able to build consistent $3$D maps of various environments, including challenging natural scenes, deploying visual loop closure over the odometry provided by inaccurate wheel encoders and the orientation by IMU. Their robust place recognition is based on Gestalt keypoint detection and description \cite{Bosse09}. Deployment of our CNN in such system would overcome the requirement of the wheel platform and the same approach would be useful for human-carried sensory gears (Pegasus, ZEB, etc.) as mentioned in the introduction.
%
%

In our previous work \cite{Velas16}, we proposed sampling the Velodyne LiDAR point clouds by \emph{Collar Line Segments (CLS)} in order to overcome data sparsity. First, the original Velodyne point cloud is split into polar bins. The line segments are randomly generated within each bin, matched by nearest neighbor search and the resulting transformation fits the matched lines into the common planes. The CLS approach was also evaluated using the KITTI dataset and achieves $7$cm error of the pairwise scan registration. Splitting into polar bins is also used in this work during for encoding the $3$D data to $2$D representation (see \Section{sec:data-encoding}).

The top ranks in KITTI Visual odometry benchmark \cite{kitti} are for last years occupied by LiDAR Odometry and Mapping (LOAM) \cite{loam} and Visual LOAM (V-LOAM) \cite{vloam} methods. Planar and edge points are detected and used to estimate the optimal transformation in two stages: fast scan-to-scan and precise scan-to-map. The map consists of keypoints found in previous LiDAR point clouds. Scan-to-scan registration enables real-time performance and only each $n$-th frame is actually registered within the map. 

The implementation was publicly released under BSD license but withdrawn after being commercialized. The original code is accessible through the documentation\footnote{\url{http://docs.ros.org/indigo/api/loam_velodyne/html/files.html}} and we used it for evaluation and comparison with our proposed solution. In our experiments, we were able to achieve superior accuracy in the estimation of the translation parameters and comparable results in the estimation of full $6$DoF (degrees of freedom) motion parameters including rotation.
In V-LOAM \cite{vloam}, the original method was improved by fusion with RGB data from omnidirectional camera and authors also prepared method which fuses LiDAR and RGB-D data \cite{dloam}.

The encoding of $3$D LiDAR data into the $2$D representation, which can be processed by convolutional neural network (CNN), were previously proposed and used in the ground segmentation \cite{cnn-gseg} and the vehicle detection \cite{cnn-vdet}. We use a similar CNN approach for quite different task of visual odometry estimation. Besides the precision and the real-time performance, our method also contributes as the illustration of general usability of CNNs for sparse LiDAR data. The key difference is the amount and the ordering of input data processed by neural network (described in next chapter and \Figure{fig:toplevel-reg-cnn}). While the previous methods \cite{cnn-vdet, cnn-gseg} process only a single frame, in order to estimate the transformation parameters precisely we process multiple frames simultaneously. 

\section{Method}

Our \emph{goal} is the estimation of transformation $\matr{T_n} = [t_n^x, t_n^y, t_n^z, r_n^x, r_n^y, r_n^z]$ representing the $6$DoF motion of a platform carrying LiDAR sensor, given the current LiDAR frame $\set{P_n}$ and $N$ previous frames $\set{P_{n-1}}, \set{P_{n-2}}, \ldots, \set{P_{n-N}}$ in form of point clouds. This can be written as a mapping $\Theta$ from the point cloud domain $\domain{P}$ to the domain of motion parameters \eqref{eq:cnn-fun} and \eqref{eq:cnn-mapping}.
Each element of the point cloud $\matr{p} \in \set{P}$ is the vector $\matr{p} = [p^x, p^y, p^z, p^r, p^i]$, where $[p^x, p^y, p^z]$ are its coordinates in the $3$D space (right, down, front) originating at the sensor position. $p^r$ is the index of the laser beam that captured this point, which is commonly referred as the ``ring'' index since the Velodyne data resembles the rings of points shown in \Figure{fig:encoding} (top, left). The measured intensity by laser beam is denoted as $p^i$.
\begin{eqnarray}
	\matr{T_n} = \Theta(\set{P_{n}}, \set{P_{n-1}}, \set{P_{n-2}}, \ldots, \set{P_{n-N}}) \label{eq:cnn-fun} \label{eq:transformation-from-clouds} \\
    \Theta: \domain{P}^{N+1} \rightarrow \domain{R}^6 \label{eq:cnn-mapping}
\end{eqnarray}

\subsection{Data encoding}\label{sec:data-encoding}

We represent the mapping $\Theta$ by convolutional neural network. Since we use sparse $3$D point clouds and convolutional neural networks are commonly designed for dense $1$D and $2$D data, we adopt previously proposed \cite{cnn-vdet, cnn-gseg} \emph{encoding} $\Epsilon$ \eqref{eq:encoding} of $3$D LiDAR data to dense matrix $\matr{M}\in\domain{M}$. These encoded data are used for actual training the neural network implementing the mapping $\tilde{\Theta}$ \meqref{eq:regression}{eq:regression-mapping}.

\begin{eqnarray}
	\matr{M} = \Epsilon(\set{P}); \quad \Epsilon: \domain{P} \rightarrow \domain{M} \label{eq:encoding}\\
    \matr{T_n} = \tilde{\Theta}(\Epsilon(\set{P_{n}}), \Epsilon(\set{P_{n-1}}), \ldots, \Epsilon(\set{P_{n-N}})) \label{eq:regression}\\
    \tilde{\Theta}: \domain{M}^{N+1} \rightarrow \domain{R}^6 \label{eq:regression-mapping}
\end{eqnarray}

Each element $\matr{m_{r,c}}$ of the matrix $\matr{M}$ encodes points of \emph{the polar bin} $\set{b_{r,c}}\subset\set{P}$ \eqref{eq:bin-encoding} as a vector of $3$ values: depth and vertical height relative to the sensor position, and the intensity of laser return \eqref{eq:3d-to-2d-2}. Since the multiple points fall into the same bin, the representative values are computed by averaging. On the other hand, if a polar bin is empty, the missing element of the resulting matrix is interpolated from its neighbourhood using linear interpolation.

\begin{eqnarray}
	\matr{m_{r,c}} = \varepsilon(\set{b_{r,c}}); \quad \varepsilon: \domain{P} \rightarrow \domain{R}^3 \label{eq:bin-encoding}\\
    \varepsilon(\set{b_{r,c}}) = \dfrac{\mathlarger{\sum}\limits_{\matr{p}\in \set{b_{r,c}}} \left[p^y, \left\lVert p^x,p^z\right\rVert_2, p^i\right]}{\vert \set{b_{r,c}} \vert} \label{eq:3d-to-2d-2}
\end{eqnarray}

\begin{figure}[t]
	\raggedleft
    \subfloat{
        \includegraphics[width=0.33\linewidth,valign=t]{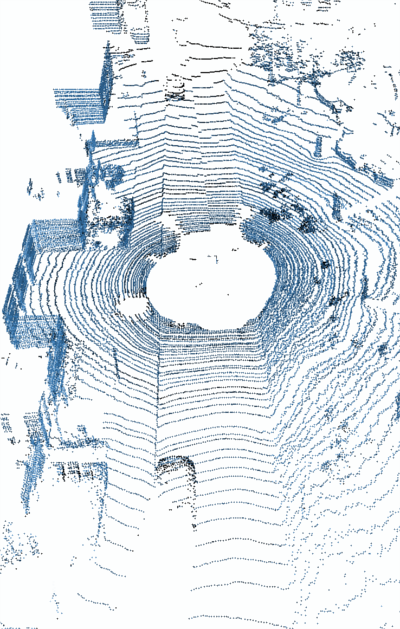}
        \includegraphics[width=0.65\linewidth,valign=t]{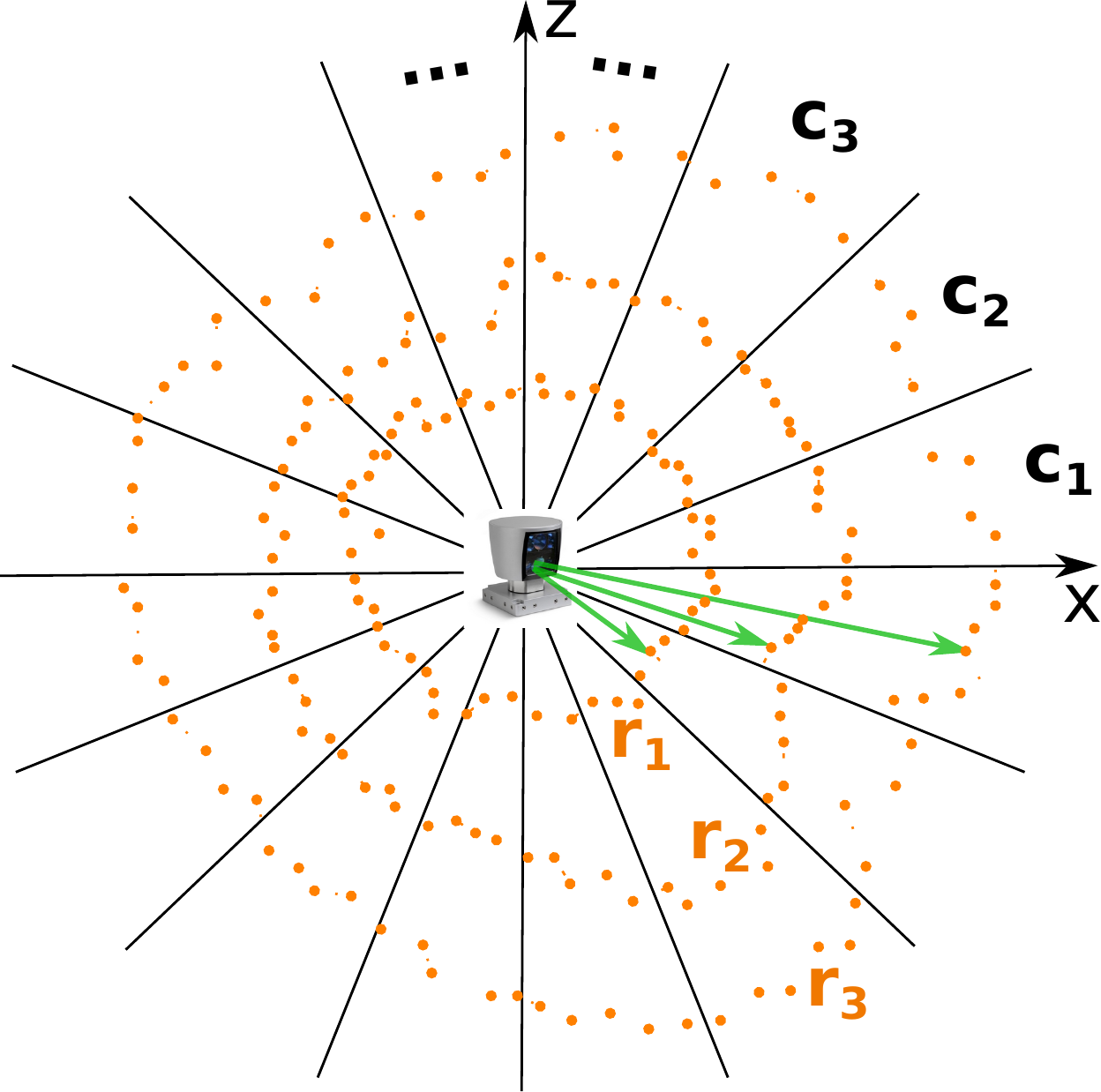}
    }

	$$\quad\quad\quad\quad\quad\quad\quad\quad\;\mathbb{\mathlarger{\mathlarger{\mathlarger{\Downarrow}}}}$$

	\vspace{-0.6em}
	\includegraphics[width=0.9\linewidth]{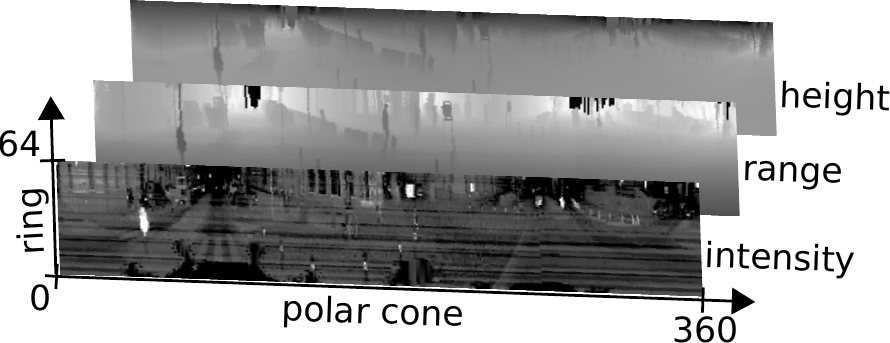}
	\caption{Transformation of the sparse Velodyne point cloud into the multi-channel dense matrix. Each row represents measurements of a single laser beam (single ring $r_1, r_2, r_3, \ldots$) done during one rotation of the sensor. Each column contains measurements of all 64 laser beams captured within the specific rotational angle interval (polar cone $c_1, c_2, c_3, \ldots$).\label{fig:encoding}}
    \vspace{-1em}
\end{figure}

The indexes $r,c$ denote both the row $(r)$ and the column $(c)$ of the encoded matrix and the polar cone $(c)$ and the ring index $(r)$ in the original point cloud (see \Figure{fig:encoding}). Dividing the point cloud into the polar bins follows same strategy as described in our previous work \cite{Velas16}. Each polar bin is identified by the polar cone $\varphi(.)$ and the ring index $p^r$.
%
%
\begin{eqnarray}
    \set{b_{r,c}} = \{ \matr{p} \in \set{P} \,\vert\, p^r = r \wedge \varphi(\matr{p}) = c \}\\
    \varphi(\matr{p}) = \left\lfloor \frac{\mbox{atan}\left(\frac{p^z}{p^x}\right) + 180^\circ}{\frac{360^\circ}{R}} \right\rfloor
\end{eqnarray}
where $R$ is horizontal angular resolution of the polar cones. In our experiments we used the resolution $R=1^\circ$ (and $0.2^\circ$ in the classification formulation described below).

\subsection{From regression to classification}

In our preliminary experiments, we trained the network $\tilde{\Theta}$ estimating full $6$DoF motion parameters. Unfortunately, such networks provided very inaccurate results. The output parameters consist of two different motion modalities -- rotation and translation $\matr{T_n} = \left[\matr{R_n} \vert \matr{t_n}\right]$ -- and it is difficult to determine (or weight) the importance of angular and positional differences in backward computation. So we decided to split the mapping into the estimation of rotation parameters $\tilde{\Theta}_{\matr{R}}$ \eqref{eq:r-by-regression} and translation $\tilde{\Theta}_{\matr{t}}$ \eqref{eq:t-by-regression}.
\begin{eqnarray}
    \matr{R_n} = \tilde{\Theta}_{\matr{R}}(\set{M_{n}}, \set{M_{n-1}}, \ldots, \set{M_{n-N}}) \label{eq:r-by-regression}\\
    \matr{t_n} = \tilde{\Theta}_{\matr{t}}(\set{M_{n}}, \set{M_{n-1}}, \ldots, \set{M_{n-N}}) \label{eq:t-by-regression}\\
    \tilde{\Theta}_{\matr{R}}: \domain{M}^{N+1} \rightarrow \domain{R}^3; \quad \tilde{\Theta}_{\matr{t}}: \domain{M}^{N+1} \rightarrow \domain{R}^3
\end{eqnarray}

\begin{figure}[t]
	\centering
    \includegraphics[width=0.9\linewidth]{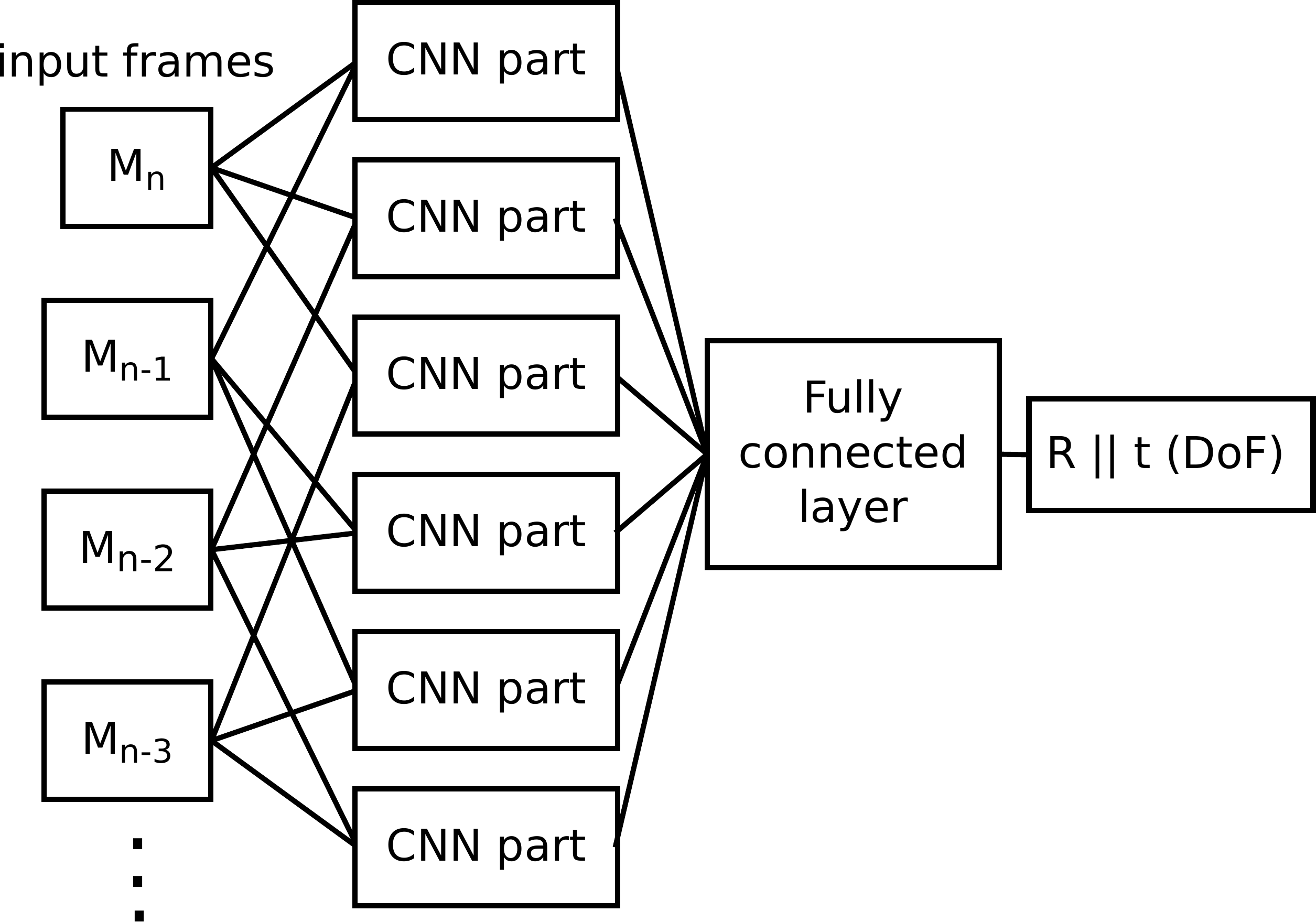}
    \caption{Topology of the network implementing $\tilde{\Theta}_{\matr{R}}$ and $\tilde{\Theta}_{\matr{t}}$. All combinations of current $\matr{M_n}$ and previous $\matr{M_{n-1}}, \matr{M_{n-2}}, \ldots$ frames ($3$ previous frames in this example) are pairwise processed by the same CNN part (see structure in \Figure{fig:cnn-part}) with shared weights. The final estimation of rotation or translation parameters is done in fully connected layer joining the outputs of CNN parts.
    \label{fig:toplevel-reg-cnn}}
    
    \vspace{1.5em}

    \includegraphics[width=0.85\linewidth]{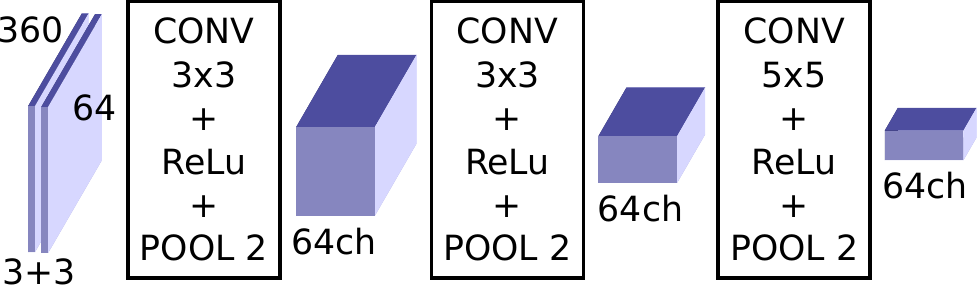}
    \caption{Topology of shared CNN component (denoted as \emph{``CNN part''} in \Figure{fig:toplevel-reg-cnn}) for processing the pairs of encoded LiDAR frames. The topology is quite shallow with small convolutional kernels, ReLu nonlinearities and max polling after each convolutional layer. The output blob size is $45 \times 8 \times 64$ ($W \times H \times Ch$).
	\label{fig:cnn-part}}
    \vspace{-1.2em}
\end{figure}

The implementation of $\tilde{\Theta}_{\matr{R}}$ and $\tilde{\Theta}_{\matr{t}}$ by convolutional neural network is shown in \Figure{fig:toplevel-reg-cnn}. We use \emph{multiple input frames} in order to improve stability and robustness of the method. Such multi-frame approach was also successfully used in our previous work \cite{Velas16} and comes from assumption, that motion parameters are similar within small time window ($0.1-0.7$s in our experiments below).

The idea behind proposed topology is the expectation that shared CNN components for pairwise frame processing will estimate the motion map across the input frame space (analogous to the optical flow in image processing). The final estimation of rotation or translation parameters is performed in the fully connected layer joining the outputs of pure CNN components.

\begin{figure}[t]
	\centering
    \includegraphics[width=0.85\linewidth]{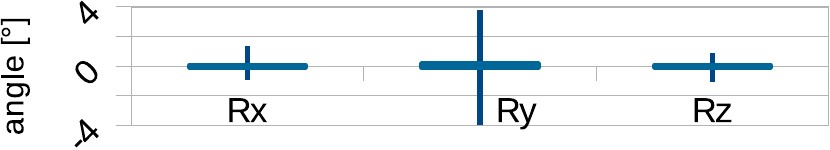}
    \caption{Rotations (min-max) around $x, y, z$ axis in training data sequences of KITTI dataset.
    \label{fig:kitti-angles}}
    
    \vspace{1em}

	\centering
    \includegraphics[width=\linewidth]{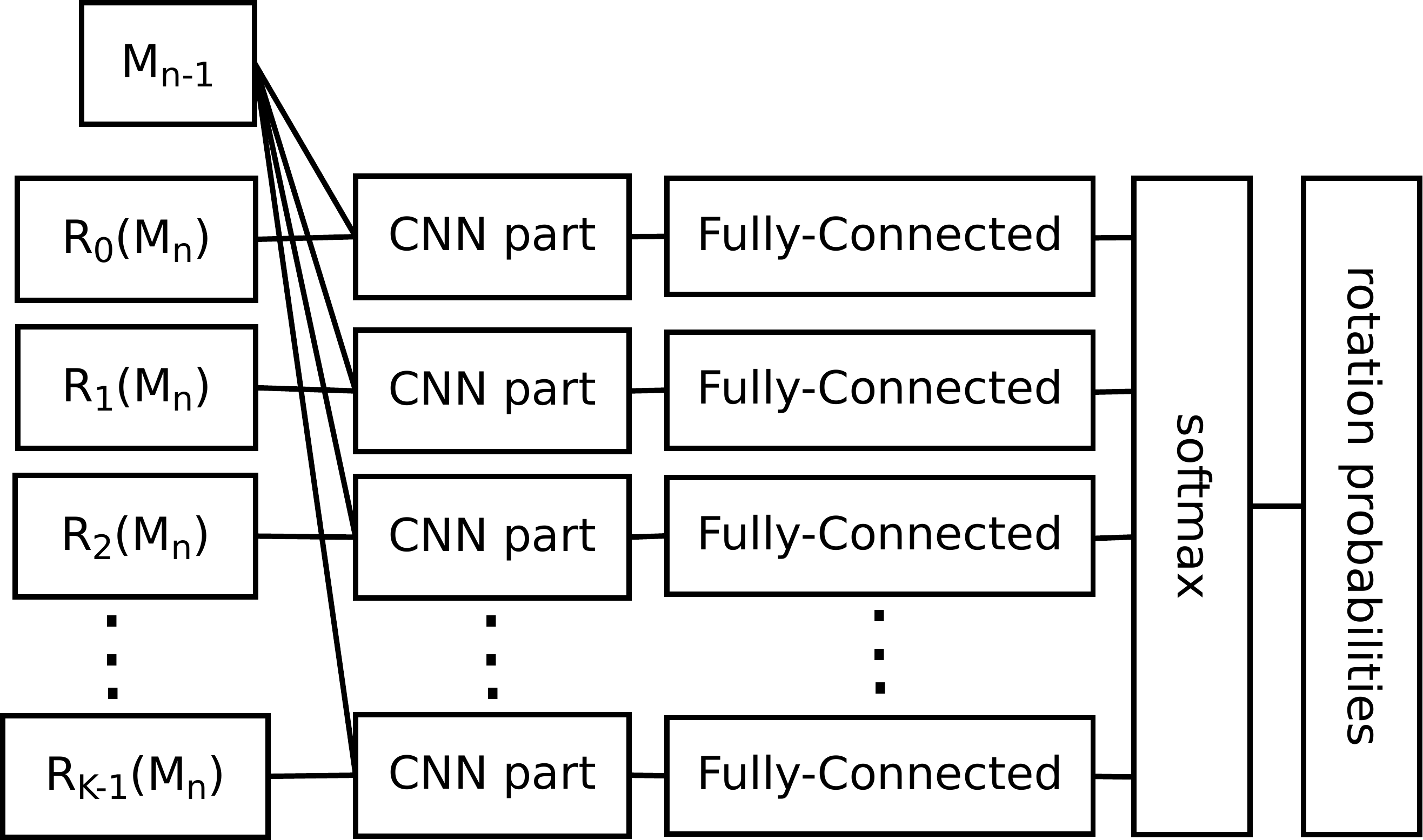}
    \caption{Modification of original topology (\Figure{fig:toplevel-reg-cnn}) for precise estimation of rotation parameters. Rotation parameter space (each axis separately) is densely sampled into $K$ rotations $R_{0}, R_{1}, \ldots, R_{K-1}$ and applied to current frame $\matr{M_n}$. CNN component and fully connected layer are trained as comparators $\Gamma$ with previous frame $\matr{M_{n-1}}$ estimating probability of given rotation. All CNN parts (structure in \Figure{fig:cnn-part-wide}) and fully connected layers share the weights of the activations.
    \label{fig:toplevel-cls-cnn}}
    
    \vspace{1.5em}

	\centering
    \includegraphics[width=\linewidth]{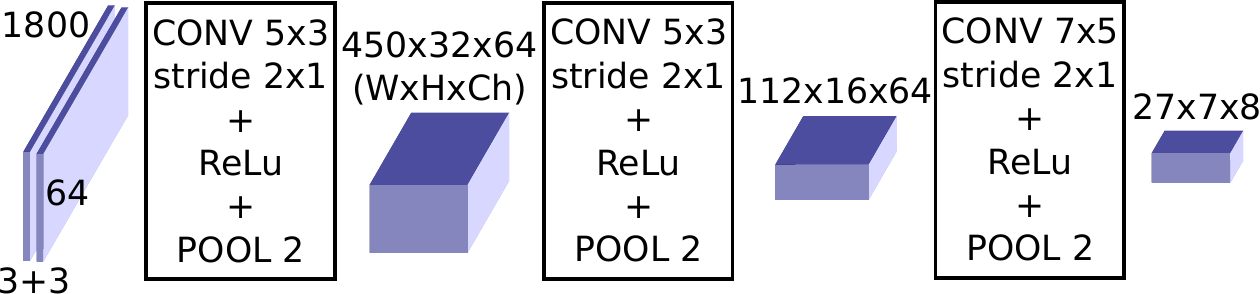}
    \caption{Modification of convolutional component for classification network. Wider input (angular resolution $R=0.2^\circ$) and wider convolution kernels with horizontal stride are used. 
    \label{fig:cnn-part-wide}}
    \vspace{-1.7em}
\end{figure}

Splitting the task of odometry estimation into two separated networks, sharing the same topology and input data, significantly improved the results -- especially the precision of translation parameters. However, precision of the predicted rotation was still insufficient. The original formulations of our goal \eqref{eq:transformation-from-clouds} can be considered as solving the \emph{regression task}. However, the space of possible rotations between consequent frames is quite small for reasonable data (distribution of rotations for KITTI dataset can be found in \Figure{fig:kitti-angles}). Such small space can be densely sampled and we can reformulate this problem to the \emph{classification task} \meqref{eq:rot-by-classification}{eq:rot-by-classification-mapping}.
\begin{eqnarray}
    R = \underset{i \in \{0, \ldots, K-1\} }{\operatorname{arg\,max}} \,\, \Gamma(R_i(\matr{M_{n}}), \matr{M_{n-1}}) \label{eq:rot-by-classification}\\
    \Gamma: \domain{M}^2 \rightarrow \domain{R} \label{eq:rot-by-classification-mapping}
\end{eqnarray}
where $R_i(\matr{M_{n}})$ represents rotation $R_i$ of the current LiDAR frame $\matr{M_{n}}$ and $\Gamma(.)$ estimates the probability of $R_i$ to be the correct rotation between the frames $\matr{M_{n}}$ and $\matr{M_{n-1}}$.

Similar approach was previously used in the task of human age estimation \cite{Rothe16}. Instead of training the CNN to estimate the age directly, the image of person is classified to be $0, 1, \ldots, 100$ years old.

The implementation of $\Gamma$ comparator by a convolutional network can be found in \Figure{fig:toplevel-cls-cnn}. In next sections, this network will be referred as \emph{classification CNN} while the original one will be referred as \emph{regression CNN}. We have also experimented with the classification-like formulation of the problem using the original CNN topology (\Figure{fig:toplevel-reg-cnn}) without sampling and applying the rotations, but this did not bring any significant improvement.

For the classification network we have experienced better results when wider input (horizontal resolution $R=0.2^\circ$) is provided to the network. This affected also properties of the convolutional component used (the CNN part), where wider convolution kernels are used with horizontal stride (see \Figure{fig:cnn-part-wide}) in order to reduce the amount of data processed by the fully connected layer.

Although the space of observed rotations is quite small (approximately $\pm1^{\circ}$ around $x$ and $z$ axis, and $\pm4^{\circ}$ for $y$ axis, see \Figure{fig:kitti-angles}), sampling densely (by fraction of degree) this subspace of $3$D rotations would result in thousands of possible rotations. Because such amount of rotations would be infeasible to process, we decided to estimate the rotation around each axis separately, so we trained $3$ CNNs implementing \eqref{eq:rot-by-classification} for rotations around $x$, $y$ and $z$ axis separately. These networks share the same topology (\Figure{fig:toplevel-cls-cnn}).

In the formulation of our classification problem \eqref{eq:rot-by-classification}, the final decision of the best rotation $R^*$ is done by max polling. Since $\Gamma$ estimates the probability of rotation angle $p(R_i)$ \eqref{eq:rot-prob}, assuming the normal distribution we can compute also maximum likelihood solution by weighted average \eqref{eq:rot-by-window}.
\begin{eqnarray}
	p(R_i) &=& \Gamma(R_i(\matr{M_n}),\matr{M_{n-1}}) \label{eq:rot-prob}\\
	R^* &=& \dfrac{\sum\limits_{i \in \set{S_W}} p(R_i).R_i}{\sum\limits_{i \in \set{S_W}} p(R_i)} \label{eq:rot-by-window} \\
    \set{S_W} &=& \underset{\set{S} = \{i_0, \ldots, i_0+W\} }{\operatorname{arg\,max}} \,\, \sum\limits_{i \in \set{S}} p(R_i) \label{eq:window-set}
\end{eqnarray}

Moreover, this estimation can done for a window of fixed size $W$ which is limited only for the highest rotation probabilities \eqref{eq:window-set}. Window of size $1$ results in max polling.

\subsection{Data processing}

For training and testing the proposed networks, we used encoded data from Velodyne LiDAR sensor. As we mentioned before, the original raw point clouds consist of $x$, $y$ and $z$ coordinates, identification of laser beam which captured given point and the value of laser intensity reading. 
The encoding into $2$D representation transforms $x$ and $z$ coordinates (horizontal plane) into the depth information and horizontal angle represented by range channel and the column index respectively in the encoded matrix. The intensity readings and $y$ coordinates are directly mapped into the matrix channels and laser beam index is represented by encoded matrix row index. This means that our encoding (besides the aggregating multiple points into the same polar bin) did not cause any information loss.

Furthermore, we use the same data normalization \eqref{eq:normalization} and rescaling as we used in our previous work \cite{cnn-gseg}. 
\begin{equation}
	\overline{h} = \frac{y^{i}}{H}; \qquad \overline{d} = \log{(d)} \label{eq:normalization}
\end{equation}
This applies only to the vertical height $h$ and depth $d$, since the intensity values are already normalized to interval $(0;1)$. We set the height normalization constant to $H=3$, since in the usual scenarios, the Velodyne (model HDL-$64$E) captures vertical slice approximately $3\mathrm{m}$ high.

In our preliminary experiments, we trained the convolutional networks without this normalization and rescaling \eqref{eq:normalization} and we also experimented with using the $3$D point coordinates as the channels of CNN input matrices. All these approaches resulted only in worse odometry precision.

\section{Experiments}

We implemented the proposed networks using \emph{Caffe}\footnote{\url{caffe.berkeleyvision.org}} deep learning framework. For training and testing, data from the KITTI odometry benchmark\footnote{\url{www.cvlibs.net/datasets/kitti/eval_odometry.php}} were used together with provided development kit for the evaluation and error estimation. The LiDAR data were collected by Velodyne HDL-$64$E sensor mounted on top of a vehicle together with IMU sensor and GPS localization unit with RTK correction signal providing precise position and orientation measurements \cite{kitti}. Velodyne spins with frequency $10$Hz providing $10$ LiDAR scans per second. The dataset consist of $11$ data sequences where ground truth is provided. We split these data to training (sequences $00$-$07$) and testing set (sequences $08$-$10$). The rest of the dataset (sequences $11$-$21$) serves for benchmarking purposes only.

The error of estimated odometry is evaluated by the development kit provided with the KITTI benchmark. The data sequences are split into subsequences of $100, 200, \ldots, 800$ frames ($10, 20, \ldots, 80$ seconds duration). The error $e_s$ of each subsequence is computed as \eqref{eq:kitti-error}.
\begin{equation}
	e_s = \dfrac{\left\lVert \matr{E_s}, \matr{C_s} \right\rVert_2}{l_s} \label{eq:kitti-error}
\end{equation}
where $\matr{E_s}$ is the expected position (from ground truth) and $\matr{C_s}$ is the estimated position of the LiDAR where the last frame of subsequence was taken with respect to the initial position (within given subsequence). The difference is divided by the length $l_s$ of the followed trajectory. The final error value is the average of errors $e_s$ across all the subsequences of all the lengths.

First, we trained and evaluated regression networks (topology described in \Figure{fig:toplevel-reg-cnn}) for direct estimation of rotation or translation parameters. The results can be found in \Table{tab:reg-results}. To determine the error of the network predicting translation or rotation motion parameters, the missing rotation or translation parameters respectively were taken from the ground truth data since the evaluation requires all $6$DoF parameters.

\begin{table}[t]
	\centering
    \def\arraystretch{1.3}
	\begin{tabular}{c||c|c|c||c|c}
    	\toprule
		{} & \textbf{CNN-t} & \textbf{CNN-R} & \textbf{CNN-Rt} & \multicolumn{2}{c}{\textbf{Forward time} [s/frame]} \\
        $\mathbf{N}$ & error & error & error & \quad\textbf{GPU}\quad\quad & \textbf{CPU} \\\hline\hline
        $1$ & $0.0184$ & $0.3794$ & $0.3827$ & 0.004 & 0.065 \\\hline
        $2$ & $0.0129$ & $0.2752$ & $0.2764$ & 0.013 & 0.194 \\\hline
        $3$ & $0.0111$ & $0.2615$ & $0.2617$ & 0.026 & 0.393 \\\hline
        $\mathbf{5}$ & $\mathbf{0.0103} $ & $0.2646$ & $0.2656$ & 0.067 & 0.987 \\\hline
        $7$ & $0.0130$ & $0.2534$ & $0.2546$ & 0.125 & 1.873 \\\hline
         \bottomrule
	\end{tabular}
	\caption{Evaluation of regression networks for different size of input data -- $N$ is the number of previous frames. The convolutional networks were used to determine the translation parameters only (column CNN-t), the rotation only (CNN-R) and both the rotation and translation (CNN-Rt) parameters for KITTI sequences $00$-$08$. Error of the estimated odometry together with the processing time of single frame (using CPU only or GPU acceleration) is presented.
    \label{tab:reg-results}}
\end{table}

\begin{table}[t]
	\centering
    \def\arraystretch{1.3}
	\begin{tabular}{c|c||c|c}
    	\toprule
		\textbf{Window size} $W$ & \textbf{Odom. error} & \textbf{Window size} & \textbf{Odom. error} \\\hline\hline
        $1$ (max polling) & $0.03573$ & $9$ & $0.03704$ \\\hline
        $\mathbf{3}$ & $\mathbf{0.03433}$ & $11$ & $0.03712$ \\\hline
        $5$ & $0.03504$ & $13$ & $0.03719$ \\\hline
        $7$ & $0.03629$ & all & $0.03719$ \\\hline
         \bottomrule
	\end{tabular}
	\caption{The impact of window size on the error of odometry, when the rotation parameters are estimated by classification strategy. Window size $W=1$ is equivalent to the max pooling, maximal likelihood solution is found also when ``\emph{all}'' probabilities are taken into the account without the window restriction.\label{tab:win-eval}}
    \vspace{-1em}
\end{table}

\begin{table*}[t]
	\centering
    \def\arraystretch{1.3}
	\begin{tabular}{r||c|c|c||c|c|c|c}
    	\toprule
		{} & \multicolumn{3}{c||}{\textbf{Translation only}} & \multicolumn{4}{c}{\textbf{Rotation and translation}}   \\\cline{2-8}
		Seq. \# & \textbf{LOAM-full} & \textbf{LOAM-online} & \textbf{CNN-regression} & \textbf{LOAM-full} & \textbf{LOAM-online} & \textbf{CNN-regression} & \textbf{CNN-classification}   \\\hline\hline
        $00$ & $0.0152$ & $0.0193$ & $0.0084$ & $0.0225$ & $0.0516$ & $0.2877$ & $0.0302$ \\\hline
        $01$ & $0.0368$ & $0.0255$ & $0.0079$ & $0.0396$ & $0.0385$ & $0.1492$ & $0.0444$ \\\hline
        $02$ & $0.0383$ & $0.0293$ & $0.0076$ & $0.0461$ & $0.0550$ & $0.2290$ & $0.0342$ \\\hline
        $03$ & $0.0120$ & $0.0117$ & $0.0166$ & $0.0191$ & $0.0294$ & $0.0648$ & $0.0494$ \\\hline
        $04$ & $0.0076$ & $0.0085$ & $0.0089$ & $0.0148$ & $0.0150$ & $0.0757$ & $0.0177$ \\\hline
        $05$ & $0.0092$ & $0.0096$ & $0.0056$ & $0.0184$ & $0.0246$ & $0.1357$ & $0.0235$ \\\hline
        $06$ & $0.0088$ & $0.0130$ & $0.0036$ & $0.0160$ & $0.0335$ & $0.0812$ & $0.0188$ \\\hline
        $07$ & $0.0137$ & $0.0155$ & $0.0077$ & $0.0192$ & $0.0380$ & $0.1308$ & $0.0177$ \\\hline\hline
        \textbf{Train average} & $0.0214$ & $0.0197$ & $0.0077$ & $0.0287$ & $0.0433$ & $0.1970$ & $0.0303$ \\\hline\hline
        $08$ & $0.0107$ & $0.0145$ & $0.0096$ & $0.0239$ & $0.0349$ & $0.2716$ & $0.0289$ \\\hline
        $09$ & $0.0368$ & $0.0380$ & $0.0098$ & $0.0322$ & $0.0430$ & $0.2373$ & $0.0494$ \\\hline
        $10$ & $0.0213$ & $0.0196$ & $0.0128$ & $0.0295$ & $0.0399$ & $0.2823$ & $0.0327$ \\\hline\hline
        \textbf{Test average} & $0.0186$ & $0.0208$ & $\mathbf{0.0102}$ & $\mathbf{0.0268}$ & $0.0376$ & $0.2655$ & $0.0343$ \\\hline
         \bottomrule
	\end{tabular}
    \caption{Comparison of the odometry estimatation precision by the proposed method and LOAM for sequences of the KITTI dataset \cite{kitti} (sequences $00-07$ were used for training the CNN, $08-10$ for testing only). LOAM was tested in the on-line mode (LOAM-online) when the time spent for single frame processing is limited to Velodyne fps ($0.1$s/frame) and in the full mode (LOAM-full) where each frame is fully registered within the map. Both the regression (CNN-regression) and the classification (CNN-classification) strategies of our method are included. When only translation parameters are estimated, our method outperforms LOAM. On the contrary, LOAM outperforms our CNN odometry when full $6$DoF motion parameters are estimated.
    \label{tab:loam-cnn-compare}}
    \vspace{-1em}
\end{table*}

Evaluation shows that proposed CNNs predict the translation (\emph{CNN-t} in \Table{tab:reg-results}) with high precision -- the best results were achieved for network taking the current and $N=5$ previous frames as the input. The results also show, that all these networks outperform LOAM (error $0.0186$, see evaluation in \Table{tab:loam-cnn-compare} for more details) in the estimation of translation parameters. On contrary, this method is unable to estimate rotations (\emph{CNN-R} and \emph{CNN-Rt}) with sufficient precision. All networks except the largest one ($N<7$) are capable of realtime performance with GPU support (GeForce GTX 770 used) and the smallest one also without any acceleration (running on i5-6500 CPU). Note: Velodyne standard framerate is $10$fps. 

We also wanted to explore, whether CNNs are capable to predict full $6$DoF motion parameters, including rotation angles with sufficient precision. Hence the classification network schema shown in \Figure{fig:toplevel-cls-cnn} was implemented and trained also using the Caffe framework. The network predicts probabilities for densely sampled rotation angles. We used sampling resolution $0.2^\circ$, what is equivalent to the horizontal angular resolution of Velodyne data in the KITTI dataset. Given the statistics from training data shown in \Figure{fig:kitti-angles}, we sampled the interval $\pm1.3^\circ$ of rotations around $x$ and $z$ axis into $13$ classes, and the interval $\pm5.6^\circ$ into $56$ classes, including approximately $30\%$ tolerance.
%
%

Since the network predicts the probabilities of given rotations, the final estimation of the rotation angle is obtained by max polling \eqref{eq:rot-by-classification} or by the window approach of maximum likelihood estimation (\ref{eq:rot-by-window},\ref{eq:window-set}). \Table{tab:win-eval} shows that optimal results are achieved when the window size $W=3$ is used.

\begin{figure*}[t]
	\centering
   	\subfloat{
    	\includegraphics[width=0.16\linewidth]{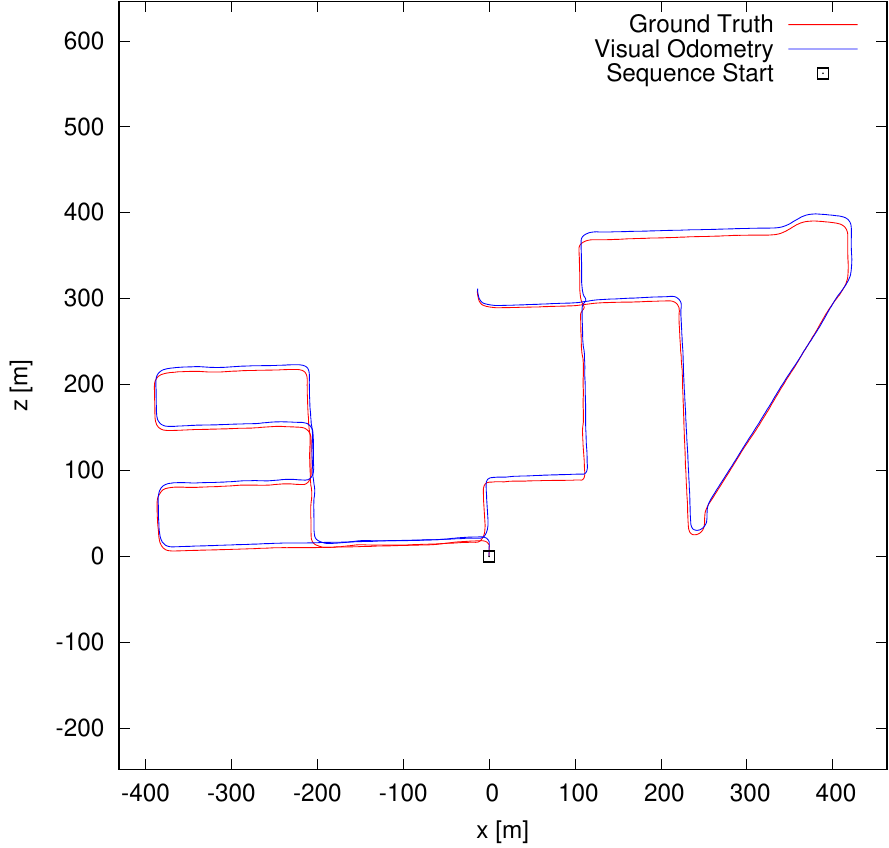}
    	\includegraphics[width=0.16\linewidth]{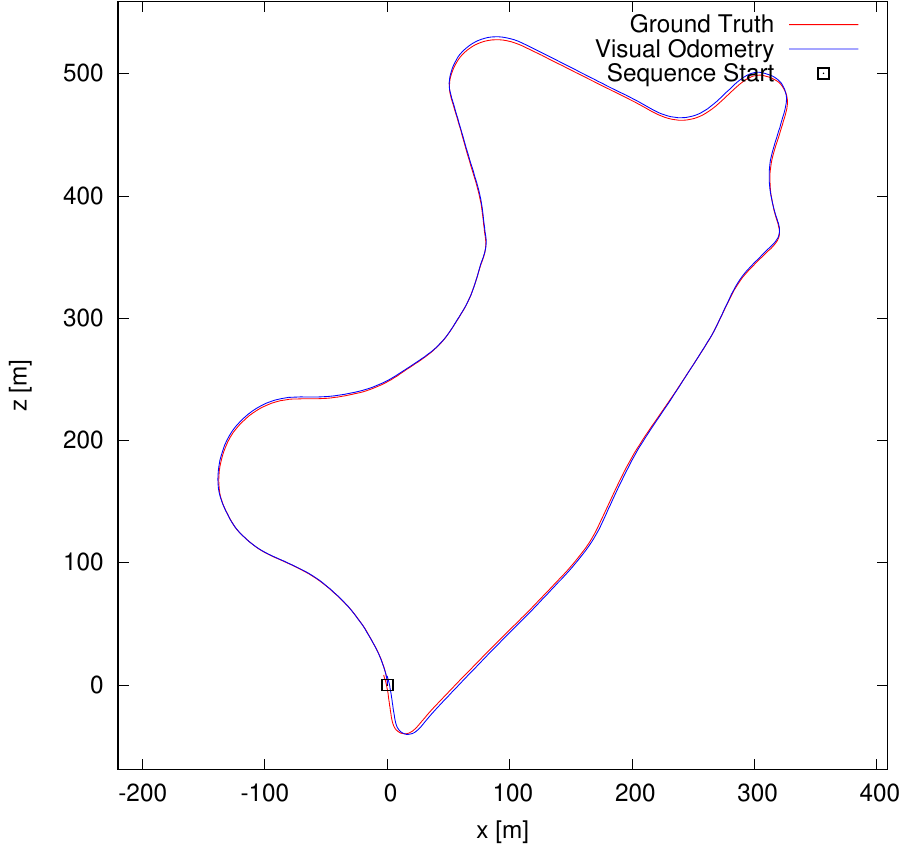}
    	\includegraphics[width=0.16\linewidth]{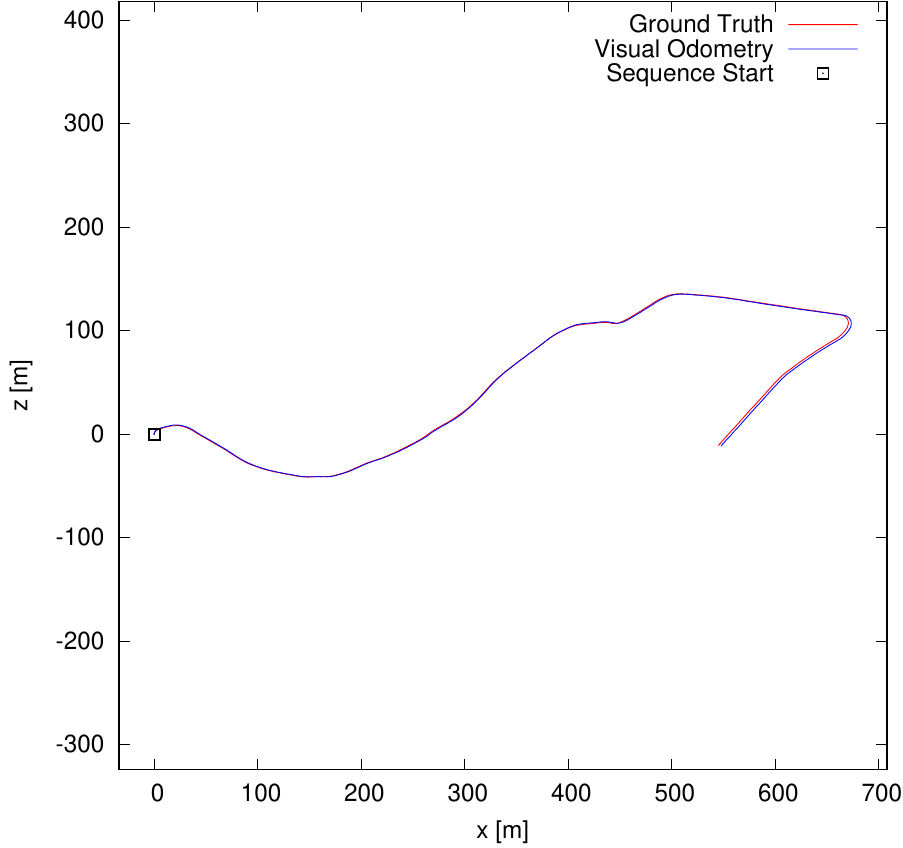}
    	\includegraphics[width=0.16\linewidth]{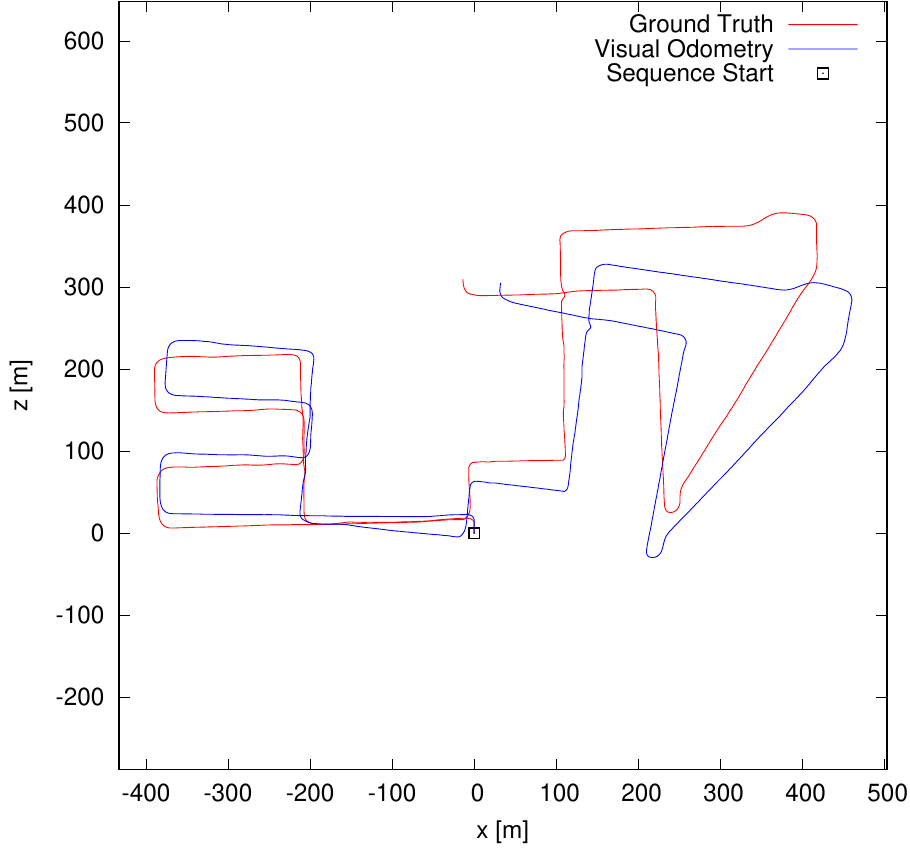}
    	\includegraphics[width=0.16\linewidth]{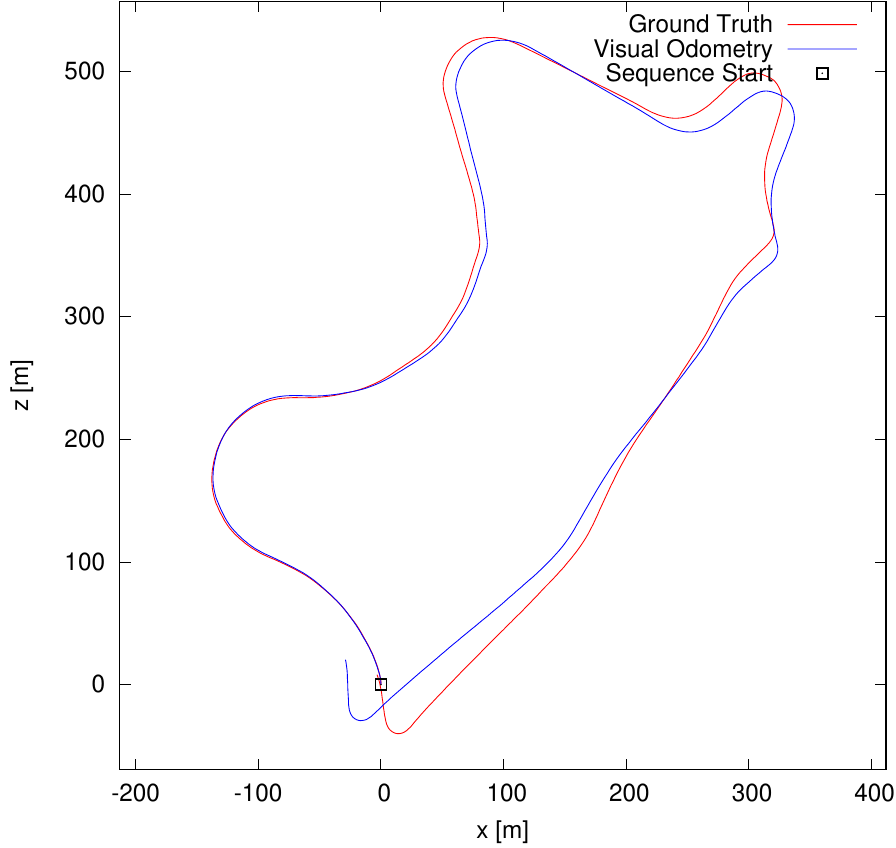}
    	\includegraphics[width=0.16\linewidth]{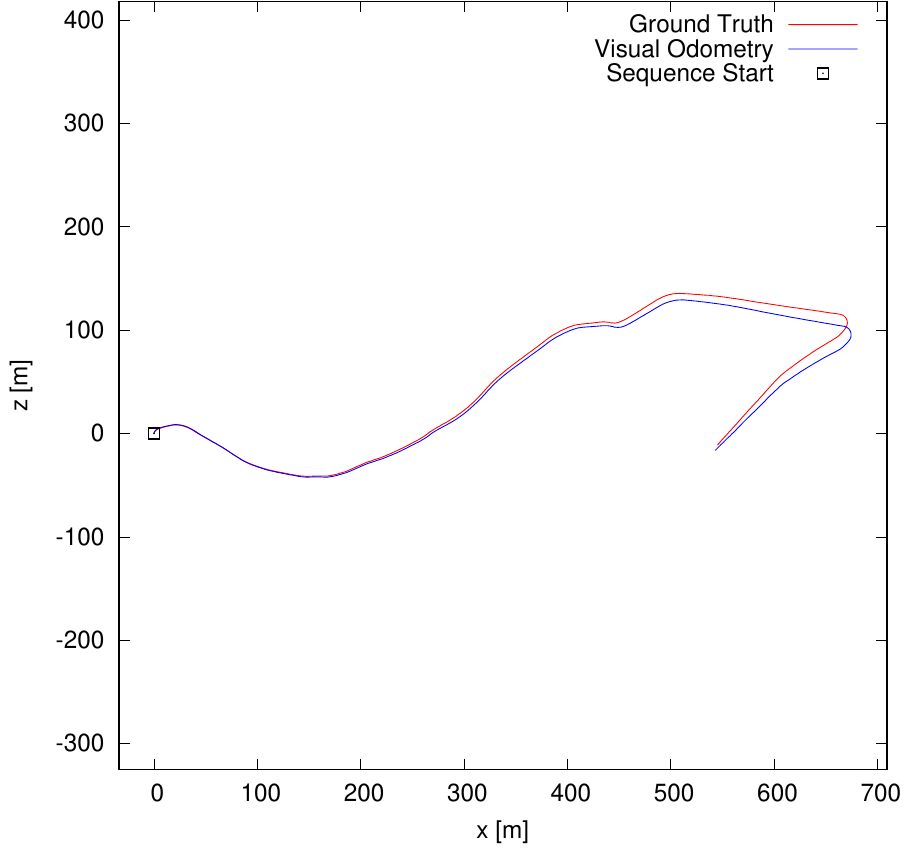}
    }

   	\subfloat{
    	\includegraphics[width=0.16\linewidth]{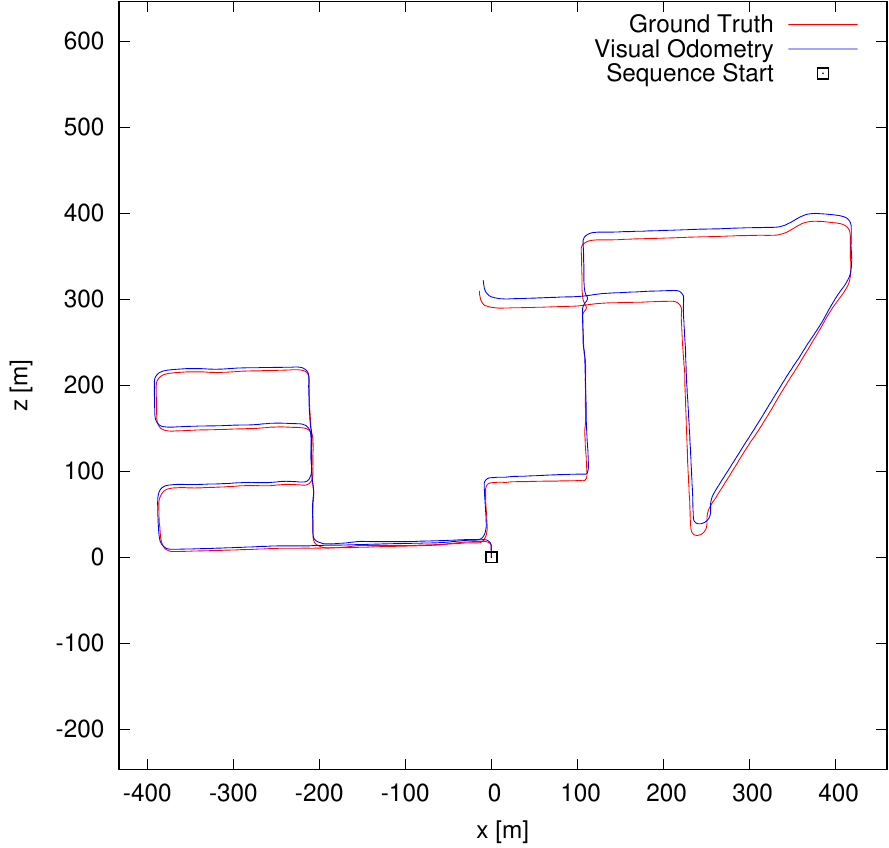}
    	\includegraphics[width=0.16\linewidth]{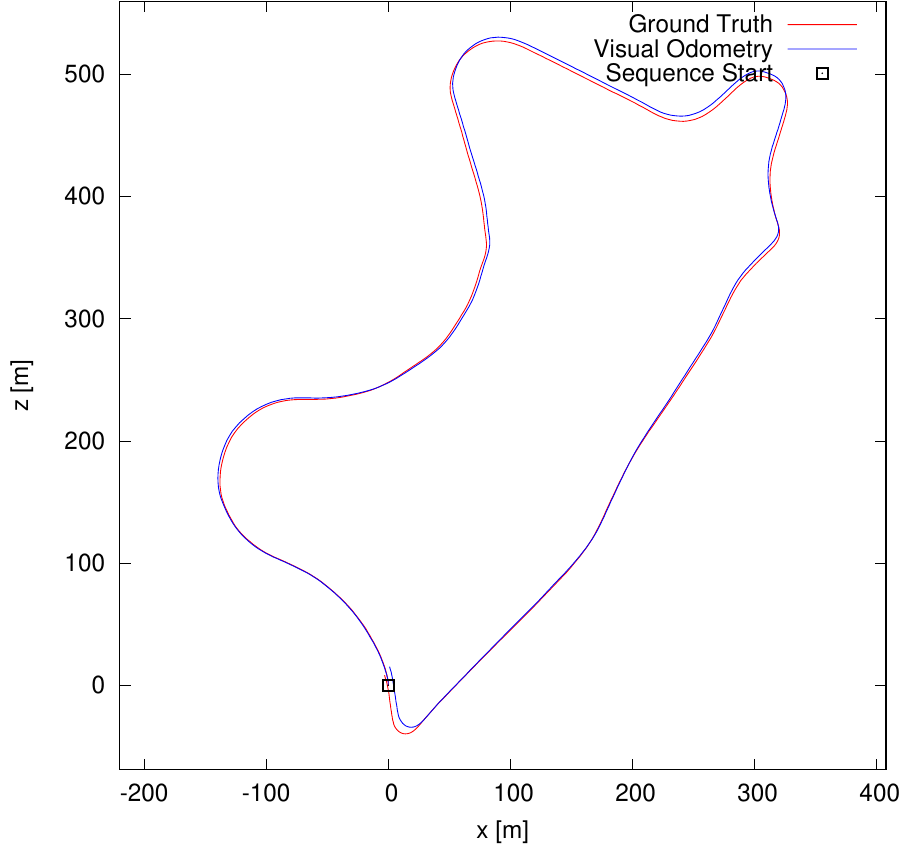}
    	\includegraphics[width=0.16\linewidth]{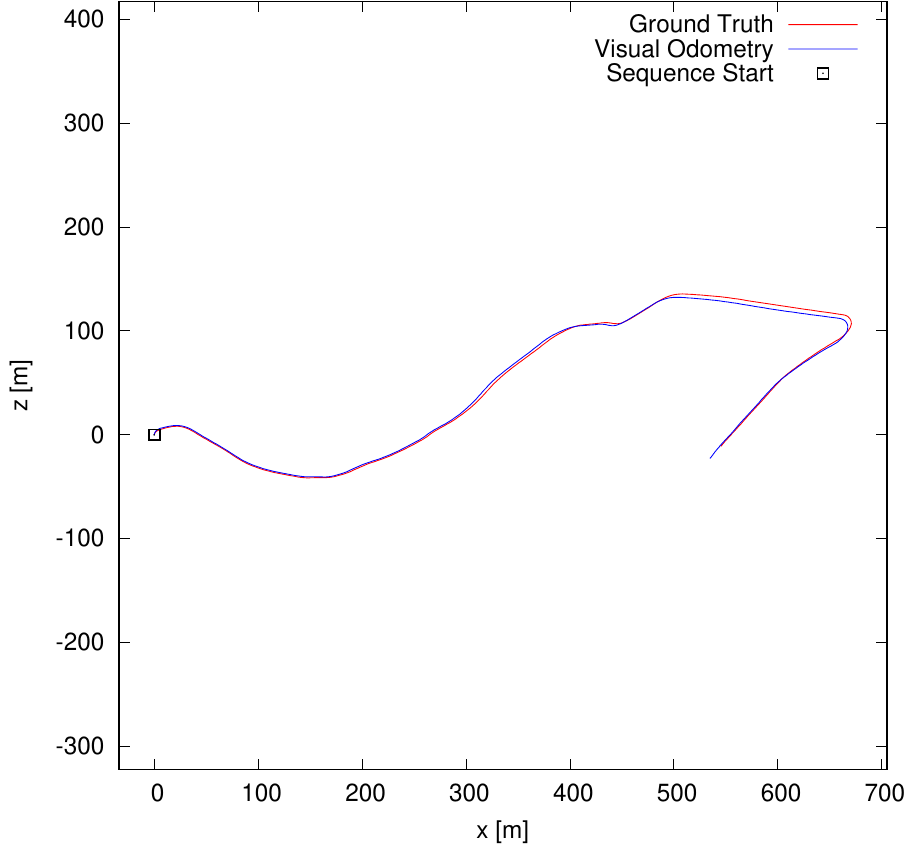}
    	\includegraphics[width=0.16\linewidth]{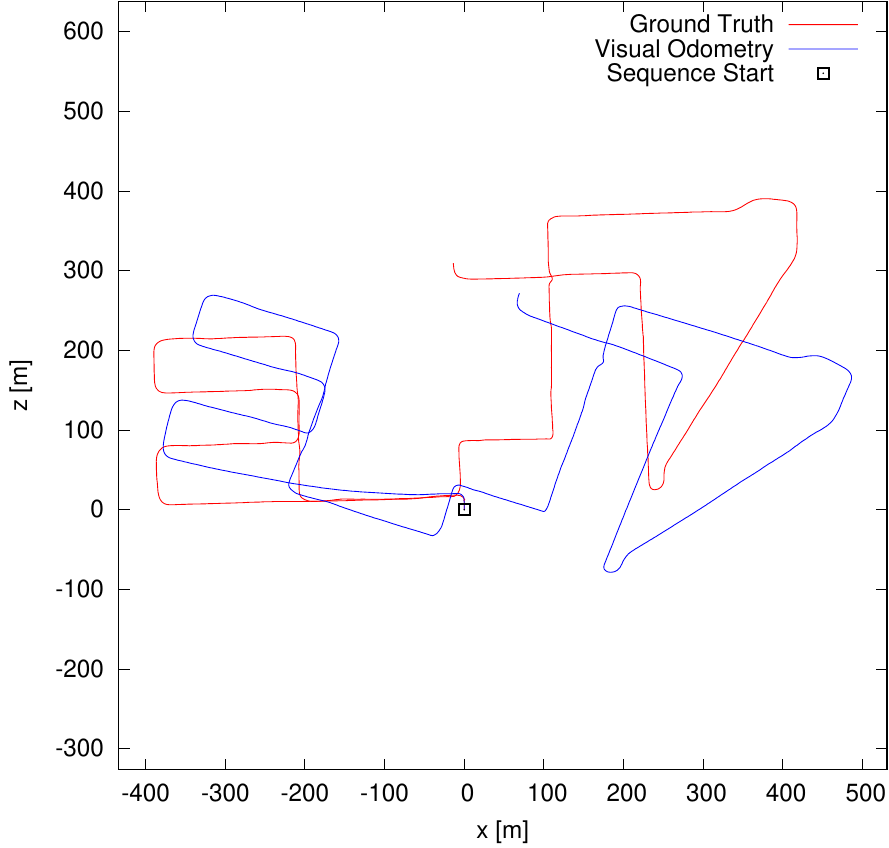}
    	\includegraphics[width=0.16\linewidth]{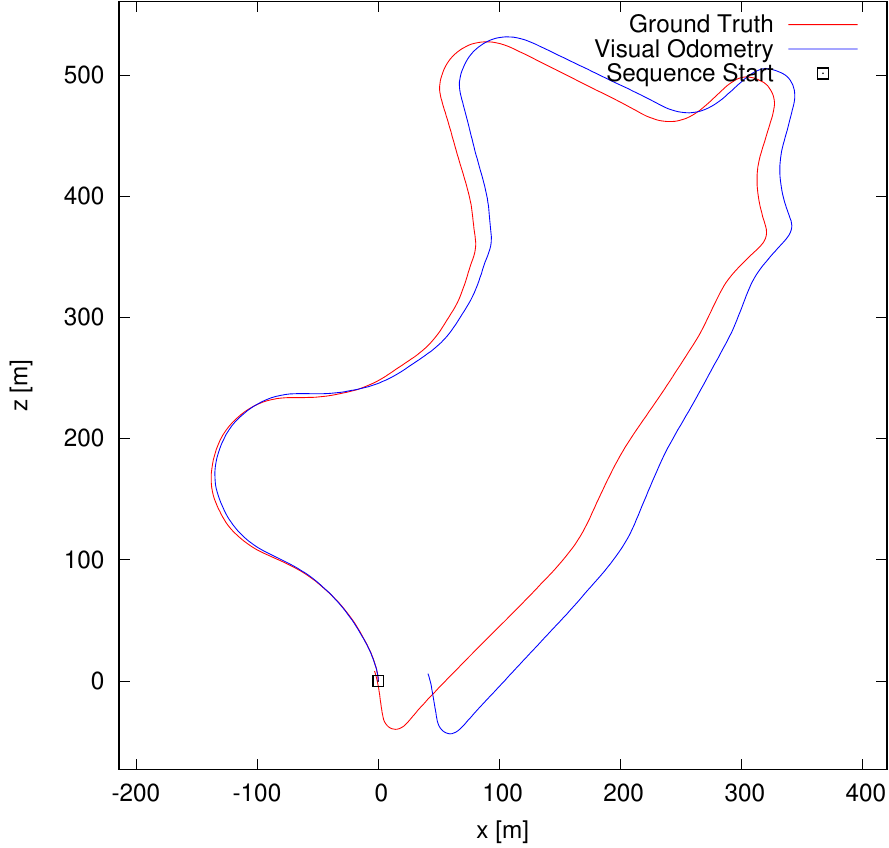}
    	\includegraphics[width=0.16\linewidth]{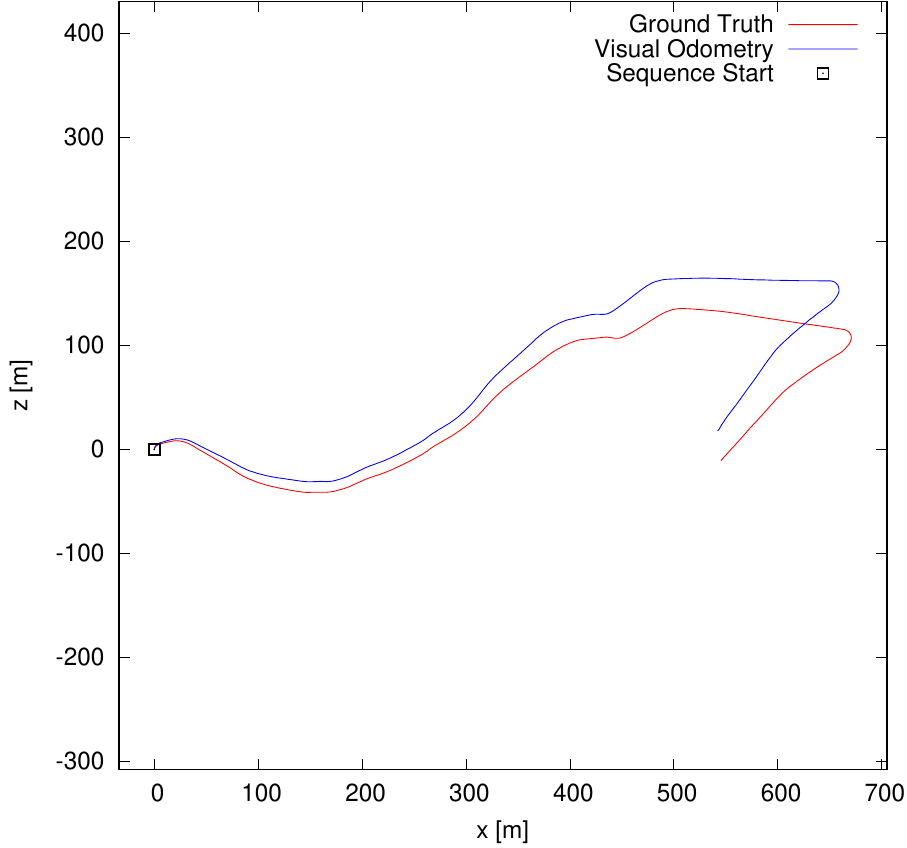}
    }
   	\caption{The example of LOAM results (top) and our CNNs (bottom row) for KITTI sequences used for testing ($08-10$). When only translation parameters are estimated (first $3$ columns), both methods achieves very good precision and the differences from ground truth (red) are barely visible. When all $6$DoF motion parameters are estimated (columns $4-6$), better performance of loam LOAM can be observed.
    \label{fig:loam-cnn-compare}}
    \vspace{-1em}
\end{figure*}

We compared our CNN approach for odometry estimation with the LOAM method \cite{loam}. We used the originally published ROS implementation (see link in \Section{sec:related-work}) with a slight modification to enable KITTI Velodyne HDL-$64$E data processing. In the original package, the input data format of Velodyne VLP-$16$ is ``hardcoded''. The results of this implementation is labeled in \Table{tab:loam-cnn-compare} as \emph{LOAM-online}, since the data are processed online in real time ($10$fps). This real-time performance is achieved by skipping the full mapping procedure (registration of the current frame against the internal map) for particular input frames.

Comparing with this original online mode of LOAM method, our CNN approach achieves better results in estimation of both translation and rotation motion parameters. However, it is important to mention, that our classification network for the orientation estimation requires $0.27$s/frame when using GPU acceleration.
%
%

The portion of skipped frames in the LOAM method depends on the input frame rate, size of input data, available computational power and affects the precision of estimated odometry. In our experiments with the KITTI dataset (on the same machine as we used for CNN experiments), $31.7\%$ of input frames is processed by the full mapping procedure.

In order to determine the full potential of the LOAM method, and for fair comparison, we made further modifications of the original implementation, so the mapping procedure runs for each input frame. Results of this method are labeled as \emph{LOAM-full} in \Table{tab:loam-cnn-compare} and, in estimation of all $6$DoF motion parameters, it outperforms our proposed CNNs. However, the prediction of translation parameters by our regression networks is still significantly more precise and faster. And the average processing time of a single frame by the LOAM-full method is $0.7$s. The visualization of estimated transformations can be found in \Figure{fig:loam-cnn-compare}.

\addtolength{\textheight}{-1cm}   

We have also submitted the results of our networks (i.e. the regression CNN estimating translational parameters only and the classification CNN estimating rotations) to the KITTI benchmark together with the outputs we achieved using the LOAM method in the online and the full mapping mode. The results are similar as in our experiments -- best performing LOAM-full achieves $3.49\%$ and our CNNs $4.59\%$ error. LOAM-online performed worse than in our experiments with error $9.21\%$. Interestingly, the error of our refactored original implementation of LOAM is more significant than errors reported for the original submission of the LOAM authors. This is probably caused by a special tuning of the method for the KITTI dataset which has been never published and authors unfortunately refused to share both the specification/implementation used and the outputs of their method with us.

\section{Conclusion}

This paper introduced novel method of odometry estimation using convolutional neural networks. As the most significant contribution, networks for very fast real-time and precise estimation of translation parameters, beyond the performance of other state of the art methods, were proposed. The precision of proposed CNNs was evaluated using the standard KITTI odometry dataset.

Proposed solution can replace the less accurate methods like odometry estimated from wheel platform encoders or GPS based solutions, when GNSS signal is not sufficient or corrections are missing (indoor, forests, etc.). Moreover, with the rotation parameters obtained from the IMU sensor, results of the mapping can be shown in a preview for online verification of the mapping procedure when the data are being collect.

We also introduced two alternative network topologies and training strategies for prediction of orientation angles, enabling complete visual odometry estimation using CNNs in a real time. Our method benefits from existing encoding of sparse LiDAR data for processing by CNNs \cite{cnn-gseg, cnn-vdet} and contributes as a proof of general usability of such a framework.

In the future work, we are going to deploy our odometry estimation approaches in real-word online $3$D LiDAR mapping solutions for both indoor and outdoor environments.

\bibliographystyle{bib/IEEEtran}
\bibliography{bib/bibliography}

\end{document}